\pdfoutput=1
\documentclass[11pt]{article}

\usepackage[preprint]{acl}

\usepackage{times}
\usepackage{latexsym}

\usepackage[T1]{fontenc}
\usepackage[utf8]{inputenc}

\usepackage{microtype}

\usepackage{inconsolata}

\usepackage{graphicx}

\usepackage{amsmath}
\usepackage{booktabs}
\usepackage[most]{tcolorbox}
\usepackage{enumitem}
\usepackage{multirow}
\usepackage{booktabs}
\usepackage{todonotes}
\usepackage{xcolor}
\usepackage{xspace}
\usepackage{highlight}
\usepackage{fontawesome}
\definecolor{set10-blue}{HTML}{4169E1}
\definecolor{google-red}{HTML}{de5246}

\definecolor{CustomBlue}{RGB}{57,83,191}
\usepackage[most]{tcolorbox}
\newtcbox{\clustertab}[1]{on line, box align=base, colback={#1},colframe={#1},size=fbox,arc=2pt,top=-1.5pt, bottom=-1.5pt, left=-1.5pt, right=-1.5pt, boxrule=0.2pt, enlarge left by=1pt}
\newcommand{\good}[1]{{\small\clustertab{gray!20}{\color{gray}$\uparrow  \mathbf{#1}$}}}
\newcommand{\bad}[1]{{\small\clustertab{gray!20}{\color{google-red}{$\downarrow \mathbf{#1}$}}}}

\newcommand{\gemmaFull}{Gemma-2 27B IT\xspace}

\title{Leveraging Domain Knowledge at Inference Time for LLM Translation: Retrieval versus Generation}
\author{Bryan Li\thanks{Work done at an internship at Google Translate Research.} \\
  University of Pennsylvania \\
  \texttt{bryanli@seas.upenn.edu} \\\And
  Jiaming Luo, Eleftheria Briakou, Colin Cherry \\
  Google \\
  \texttt{\{jmluo,ebriakou,colincherry\}@google.com} \\}

\begin{document}
\maketitle
\begin{abstract}
% abstract 3
% Large language models (LLMs) have been increasingly adopted for machine translation (MT), and their ability to generalize from knowledge provided in prompts makes them appealing for translating specialist domains. But how should this domain-specific knowledge be represented? Prior LLM MT work has considered the strategies of few-shot \textit{demonstrations} and \textit{terminologies}. We can also consider the knowledge source; while earlier efforts used \textit{retrieval} from external tools, recent efforts have shown some success with \textit{generation} of knowledge from an LLM's own parametric memory.
% % though only for general-domain MT.

% In this work, we study domain-adapted MT with LLMs with a careful prompting setup that decouples the representations of knowledge into \textit{sources} and \textit{strategies}. 
% On strategies, demonstrations (demos) consistently outperform terminology. On sources, retrieval outperforms generation. Generating demos can viably improve weaker models, but only to the level of static domain-specific demos.
% Given the success of demonstrations, we perform detailed analyses to understand their value. We find that domain-specificity is particularly important, and that the popular multi-domain benchmark is testing adaptation to a particular writing style more so than to a specific domain.

While large language models (LLMs) have been increasingly adopted for machine translation (MT), their performance for specialist domains such as medicine and law remains an open challenge. Prior work has shown that LLMs can be \textit{domain-adapted} at test-time by \textit{retrieving} targeted few-shot demonstrations or terminologies for inclusion in the prompt. Meanwhile, for \textit{general-purpose} LLM MT, recent studies have found some success in \textit{generating} similarly useful domain knowledge from an LLM itself, prior to translation. Our work studies domain-adapted MT with LLMs through a careful prompting setup, finding that demonstrations consistently outperform terminology, and retrieval consistently outperforms generation. We find that generating demonstrations with weaker models can close the gap with larger model's zero-shot performance. Given the effectiveness of demonstrations, we perform detailed analyses to understand their value. We find that domain-specificity is particularly important, and that the popular multi-domain benchmark is testing adaptation to a particular writing style more so than to a specific domain.
% %% End Abstract v2.1
\end{abstract}

\section{Introduction}

% Introduce and motivate the problem setting.
Large language models (LLMs) have emerged as the next major paradigm for machine translation (MT), with increasing use in both industrial and academic settings. These models are exciting not only for their strong base (or \textit{zero-shot}) translation capabilities, but also for their ability to be modified at inference time through alternate prompts~\cite{kojima2022large,kong-etal-2024-better}, in-context learning~\cite{brown2020language} and the use of intermediate reasoning~\cite{wei2024chain}. 

% Discuss our prime related works.
This flexibility is particularly exciting for adapting LLMs to translate specialist domains, such as legal or medical texts. In the statistical and neural MT eras, domain adaptation would typically take the form of an expensive continued training procedure on in-domain data~\cite{freitag2016fastdomainadaptationneural,thompson-etal-2019-overcoming}. With LLMs, there is the promise of simple adaptation at inference time. 

One promising technique is the retrieval of instance-specific demonstrations of translation from a bitext datastore for few-shot in-context learning, which has shown large improvements for domain-adapted MT~\cite{agrawal-etal-2023-context,tan-etal-2024-narrowing}, rivaling the performance of specialized nearest-neighbor MT systems~\cite{khandelwal2021nearest}. LLMs have also been shown to make good use of bilingual terminology dictionaries for lexical translation hints~\cite{ghazvininejad2023dictionary,lu2023chain,moslem-etal-2023-adaptive}.  

Intriguingly, two recent approaches have forgone external resources in favor of querying an LLM to generate useful knowledge from its internal memory. First, the MAPS approach issues LLM queries for topics, terminology, and demonstrations based on the source text~\cite{he-etal-2024-exploring}. Their terminology and demonstrations mirror the knowledge sourced from retrieval steps in earlier work. The idea is that the LLM has seen relevant information during pre-training, and would benefit from explicitly surfacing it before translation. Second, the step-by-step MT approach queries its LLM to translate and discuss idiomatic phrases before performing a complete translation~\cite{briakou2024sbys}. However, both these works only consider the general domain. This inspires us to consider the applicability of internal memory approaches to domain adaptation, for which relevant external resources may be more difficult to obtain.

\begin{figure*}[t!]
    \centering
    \includegraphics[width=\textwidth]{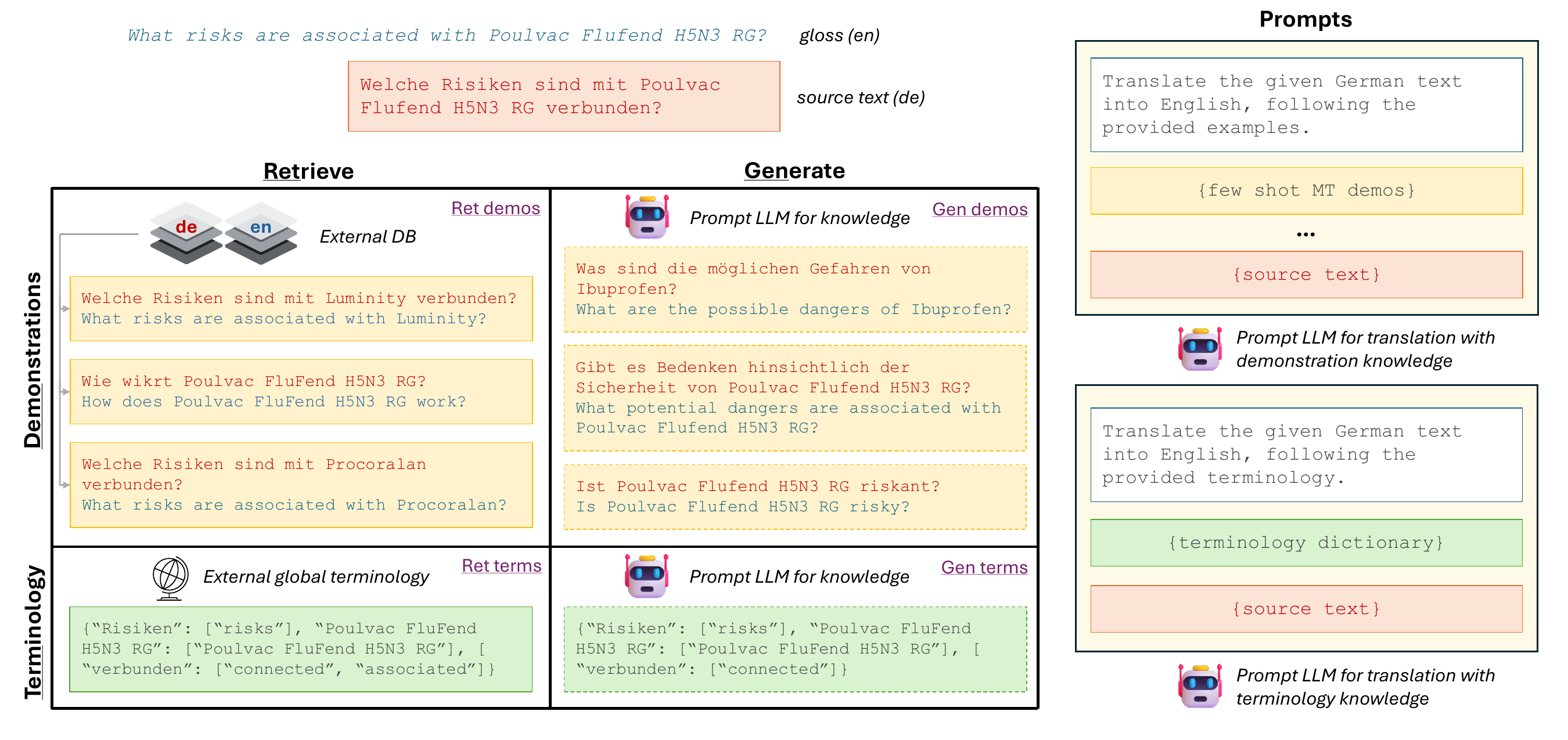}
    \caption{Illustration of the main MT settings, for an example source text in German. The two knowledge strategies are demonstrations vs. terminology; the two sources are retrieval vs. generation. This gives 4 settings for comparison. Within a strategy, we use the same prompts, varying only the provided information.}
    \label{fig:compare_settings}
\end{figure*}

% Introduce our specific study, make reference to Figure 1.
In this work, we study the effectiveness of different representations of domain-specific knowledge, in \textit{strategies} -- external retrieval vs. internal generation -- and \textit{sources} -- translation demonstrations and bilingual terminology. We consider three domains (law, medical, and Koran) from the commonly-used multi-domain dataset~\cite{aharoni-goldberg-2020-unsupervised}, and experiment with two LLMs (Gemini 1.5 and Gemma-2).
Our study addresses three main research questions:

%
% List out contributions.
%
% External knowledge further boosts the the strong language capabilities of LLMs by providing additional task and example-specific guidance. In the MT world, two such knowledge modes are \textit{demonstration retrieval} and \textit{terminology lookup}, and both have been time-tested and effective. In this work, we study how these two paradigms inter-relate, and whether we can simulate the external knowledge calls with an LLM's own parametric memory.
%
% We first follow prior work, in studying terminology and demonstration independently, in the external retrieval setting. Of course, incorporating external knowledge is effective in improving translation quality. The rest of 
\noindent \textbf{RQ1}. For improving domain-adapted MT, how viable is generation from an LLM's parametric memory compared to retrieval from external resources? 

\noindent \textbf{RQ2}. Likewise, how does adapting MT with demonstrations compare with terminologies, regardless of their source? 

\noindent \textbf{RQ3.} Given the effectiveness of demonstrations, can we attribute which of their aspects contribute the most for both retrieval and generation?

We discuss knowledge sourced from retrieval in \S\ref{sec:external} and from generation in \S\ref{sec:internal}.
Comparisons between terminology and demonstrations are enabled by our use of a silver terminology dictionary, built by LLM analysis of the same bitext used as the datastore of demonstrations (\S\ref{sec:term_induction}). This allows us to study demonstrations and terminology as alternate views into the same base data in the retrieval setting. We address RQ1 and RQ2 with the results in \S\ref{sec:results}. We explore RQ3 through several analyses in \S\ref{sec:analysis}; the main takeaways are that retrieved demonstrations mainly provide hints of target style rather than terminology, and that generated ones can viably boost performance, albeit to the same level as static domain-specific demonstrations.

% Section(s) that explain(s) how we come to design our experiments.
% \input{sections/retrieval_and_generation}
\section{Leveraging Domain Knowledge}
\label{sec:domain_knowledge}
Comparisons between representations of domain knowledge are enabled by our careful prompting setup which decouples the \textit{source} and \textit{strategy}, as sketched in Figure~\ref{fig:compare_settings}. Bilingual terms, whether retrieved externally or generated by an auxiliary LLM call, feed into the same translate-with-terms prompt, and likewise for  demonstrations. 
On \textit{sources}, retrieval leverages resources such as datastores and dictionaries, while generation elicits information from an LLM's own parametric memory.  On \textit{strategies}, demonstrations provide source-target example pairs, whereas terminology focuses on domain-specific lexical items. This section details the integration of these strategies and knowledge sources within our experimental framework for domain adaptation of LLM MT.
% Our experimental design is based on two dominant sources for obtaining supplementary information for LLMs: \emph{retrieval} and \emph{generation}. These sources are further coupled with two established strategies of representing domain knowledge: \emph{demonstrations} and \emph{terminology}.

\subsection{Knowledge from Retrieval}
\label{sec:external}
We describe two successful approaches to retrieve domain knowledge from external resources: \textit{demonstration retrieval} and \textit{terminology lookup}. %Demonstrations are source-target text pairs related to a given source text, while terminologies are dictionaries which map a source term to a target term, thus providing a pre-defined translation.  
The two related approaches operate in different fashions. Demonstration retrieval has the model \textit{implicitly} learn from the characteristics of the exemplars, both style and terminology. Terminology lookup has the model \textit{explicitly} see which source terms are important and also how to translate them.

\paragraph{Resource requirements} These methods, while effective, are expensive, as they  require the existence of high-quality and domain-specific resources. The former requires a large pool of bitext demonstrations, and the latter requires the creation of a term-rich bilingual dictionary.

% Other studi studies have largely focused on post-translation refinement, leveraging feedback from automated metrics~\cite{feng-etal-2024-improving,xu-etal-2024-llmrefine}

\subsubsection{Demonstration Retrieval}
\label{sec:demonstration_retrieval}
% In-context learning (ICL) has proved to be a dominant paradigm in interfacing with LLMs, given its flexibility and performance~\cite{brown2020language, patelBidirectional2023}. This involves conditioning a LLMs' responses to a given task instance by providing exemplars in the prompt. 
Demonstrations are provided as exemplars in the prompt to facilitate in-context learning (ICL)~\cite{brown2020language, patelBidirectional2023}. These exemplars can either be \textit{static}, the same across all instances, or \textit{instance-specific}, in which different exemplars are retrieved for each instance to provide specific guidance and hints.

The typical setup for demonstration retrieval for MT is as follows. Given a source text, we find \textit{k} closest source-side matches in an external datastore, using some similarity metric, such as BM25 or cosine similarity of embedding vectors. Then, we include in the LLM prompt these \textit{k} source texts paired with the gold target translations.
%We will refer to this approach as retrieving demonstrations (ret demos), and note it is synonymous with few-shot MT and ICL retrieval.
\paragraph{Prior work}

The use of demonstrations has a long history in MT, with some of the oldest data-driven approaches to MT having as their first step finding the most relevant examples from a bilingual translation memory. This idea has been used for computer-aided translation~\cite{yamada2011effect}, example-based MT~\cite{sommers1999review,lepage2005purest} and statistical MT~\cite{koehn-senellart-2010-convergence}.

% With the advent of LLMs, of course, renewed interest has been placed on how demo retrieval can improve MT with LLMs.
Several recent papers have studied what constitutes effective demonstration retrieval for MT with LLMs, with a particular focus on the multi-domain dataset. \citet{agrawal-etal-2023-context} found a strong baseline to be example-specific BM25 retrieval of bitexts, which can be strengthened further by re-ranking for lexical diversity.
\newcite{tan-etal-2024-narrowing} use a much larger LLM, and show that BM25 retrieval of target sentences alone can compare favorably with both sides of bitexts.
%underscore the importance of demonstrations' source proximity and lexical overlap. Using a small LLM (with low zero-shot scores), they find that their retrieve-then-rerank approach greatly boosts BLEU over a BM25 approach. 
Conversely, in the general-domain, 
researchers have found that a demonstration's quality matters more than its proximity~\cite{vilar-etal-2023-prompting,zhangPromptingLargeLanguage2023}.

\paragraph{Our Setup} For our few-shot implementation, we design a simple prompt (shown in Figure~\ref{fig:prompt_demos}). We use k=3 exemplars,
\footnote{k=3 is used to align with the 3 exemplars generated in a later experiment. COMET for k=3 and k=10 differ by \texttildelow 0.3.} and retrieve using the BM25 metric.
Our datastore, derived from the train split of multi-domain, has 16,775 demonstrations for Koran, 234,352 for medical, and 464,295 for law.
%\footnote{We follow \citet{tan-etal-2024-narrowing}, who find that for a similarly powerful LLM as the ones we study, BM25 is enough and more advanced similarity metrics are less necessary (as they are for weaker LLMs). }

\subsubsection{Terminology Lookup}
\label{sec:term_lookup}
Intuitively, one of the major challenges when translating in a specialist domain is the adaptation to domain-specific terminology. Especially in high-stakes legal, medical or business domains, precision of terminology can be crucial. Bilingual dictionaries of terminology are therefore likely sources of useful external knowledge to add into an MT system. These resources can be easier to construct than the large datastore of translation demonstrations needed in \S\ref{sec:demonstration_retrieval}. In fact, the construction of a clear terminology may very well be a prerequisite to creating human translations.

%LLM MT systems work well in the general domain, but often struggle in domains with specialized vocabulary. Terms from these domains, such as medicine and law, often require careful precision to ensure proper communication. \textbf{MT with terminologies} thus includes pre-defined terminologies in the task formulation. MT with terminologies is especially applicable to domain adaptation settings, given its aforementioned explicitness. 
%In this work, we refer to terminology lookup as ret terms, using retrieval as a synonym for lookup.

%An effective MT with terminologies system satisfies three properties: 1) general translation quality, 2) translation quality of specific terms (term success rate), and 3) translation consistency. These are the evaluation criteria for the shared tasks on terminologies hosted at recent WMT conferences~\cite{semenov-etal-2023-findings, alam-etal-2021-findings}.
%\footnote{While WMTs each have multiple shared tasks, we use WMT21 and WMT23 to refer to only the terminology task.} 
% WMT21 considers a single domain, COVID-19, and translating from English to 5 other languages, while WMT23 considers 3 domains: web novel (zh-en), NLP abstracts (en-cs), and medical abstracts (de-en). Of the WMT23 submissions, LLM-based systems tended to have higher trans

\paragraph{Prior Work} 
Improving translations with terminologies has been heavily studied.
In the statistical and neural eras, solutions could take the form of incorporating dictionaries into training~\cite{wu-etal-2008-domain}, or controllable MT systems that respect example-specific terminology constraints included in the input~\cite{post-vilar-2018-fast,wang-etal-2022-template}. More recently, terminology constraints have been studied at two WMT shared tasks~\cite{alam-etal-2021-findings,semenov-etal-2023-findings}. These approaches illustrate two different motivations for the use of terminology dictionaries in MT: the dictionary can be viewed as a useful source of domain-specific information, or as a set of constraints that must be followed consistently. Our work aligns with the former motivation, viewing bilingual terminologies only as hints to improve overall quality.

With the advent of LLMs, terminologies can be included in the prompt, with additional instructions on their usage. 
Most LLMs follow these instructions easily, as shown at the WMT23 shared task on terminology~\cite{semenov-etal-2023-findings}.
For example, \citet{moslem-etal-2023-adaptive} find that for the COVID-19 domain, a prompt using retrieved terminologies significantly boosts term success rate and also improves human evaluation scores. Other works have explored how to more effectively format dictionaries~\cite{lu2023chain,ghazvininejad2023dictionary}.

% An important consideration for terminologies is for what use-case they were created. In the WMT21 shared task, there is a \textit{global-level terminology} which was created externally, then applied to the test source texts (by looking up which source terms appear in globally, and appending that source-target(s) entry).  In contrast, for WMT23, terminologies are created directly from the test source texts; this means they are internally-defined.

% In this work, we follow the WMT21 approach, which defines terminologies as an external resource. This has two advantages: this is a more realistic MT setting, since for any arbitrary source text, we can get its relevant terms; and in this way, we can fairly compare between retrieving over demonstrations, vs. retrieving over terms. For a further discussion on characterization of terminologies, we refer readers to Appendix~\ref{sec:terms}. Therefore, we will use \textit{terminology lookup} and \textit{terminology retrieval} interchangeably.

\paragraph{Our Setup} Since the multi-domain dataset does not have a provided domain-specific terminology, we derive one from the multi-domain training set, as described in \S\ref{sec:term_induction}. Keeping with our theme of providing hints rather than constraints, the dictionary gives a list of possible translations for each source term, each licensed by at least one example in the training set. 

With this dictionary in place, we look up terms by exact lexical match to the source text currently being translated, and include any matches in our prompt for translation with terminology (shown in Figure~\ref{fig:prompt_terms}). The LLM is instructed to pick the most appropriate translation among the choices, given the source. Note that the translation prompt also includes three domain-specific examples of how to translate with terminologies.
% We carefully considered existing domain MT datasets with terminology annotations. We found that neither the terminologies from WMT21 or WMT23 were suitable for our use-case -- the latter because they is internally defined, and the former because of data leakage concerns. These concerns are further discussed in Appendix~\ref{sec:other_ds}.

% We thus extend the multi-domain dataset with annotations on terminologies, which are induced from the train bitexts. We document our approach, which is an LLM prompting-based approach adapted from prior work in \S\ref{sec:term_induction}.
% Furthermore, our extended annotations enable the careful analytical approach we take ahead, as we have a single dataset wherein we can compare the effects of using terms or demonstrations, and disentangle their individual contributions.

\subsection{Knowledge from Generation}
\label{sec:internal}
%It is evident that retrieval over external resources can boost MT performance. This  is satisfying for other knowledge-intensive NLP tasks, wherein the pretraining data may not contain the exact information required, or not in the exact output format. However, machine translation is among the tasks that modern LLMs are explicitly trained for. Furthermore, given the huge size of contemporary pretraining corpora, they do contain some amount of specialist domain data.
% Therefore, we look to study whether we can improve translations by \textit{simulating} external resources by using LLM's parametric memory. After all, \textit{retrieving} necessitates the costly collects of high-quality, and domain-relevant, external resources. 
While external knowledge retrieval demonstrably benefits knowledge-intensive NLP tasks, whether it is truly necessary for domain-adapted MT still warrants investigation, given that LLMs are explicitly trained on massive corpora including texts from specialist domains. Therefore, we investigate whether leveraging LLMs' internal parametric memory can offer comparable benefits, and thus circumvent the costly acquisition and curation of external resources. This approach effectively simulates external retrieval by prompting the LLM to generate relevant information.
% Recall that several papers found that it is not enough to just have a large bitext -- they should be high-quality, and lexically and semantically relevant for the most effective boost to MT scores~\cite{vilar-etal-2023-prompting, agrawal-etal-2023-context,zhangPromptingLargeLanguage2023}.
\paragraph{Resource requirements} By design, the generation setting requires almost no external resources. The approaches discussed below only required us to manually create a handful of static exemplars for each subtask, which are used for all of its prompts.

\label{sec:simulate}
\subsubsection{Prior Work}
Prior work has explored several methods to leverage an LLM's parametric knowledge to improve MT quality, either post-translation, or pre-translation. 
% \citet{chen-etal-2024-iterative} propose an iterative process to elicit from an LLM successively refined translations. 
Most relevant to our work are two studies which operate at the pre-translation stage.

\citet{he-etal-2024-exploring} propose a human-like translation process, where they separately prompt LLMs for 3 aspects related to a source text (demonstrations, topics, and terms). Directly using these generated knowledge pieces in another LLM interaction is insufficient, and so they rely on an external quality estimation (QE) method to select among candidates, improving general domain MT quality. Our generation setting also use demonstrations and terms, but without any external feedback from QE.

\citet{briakou2024sbys} propose a method to model the LLM translation process step-by-step. Their 2-step approach has an LLM first perform research on idiomatic expressions, then perform the full translation. For document-level MT datasets, they find this consistently outperforms zero-shot MT.

\subsubsection{Demonstration Generation}
We author a prompt to generate demonstrations (Figure~\ref{fig:prompt_demo_gen}). For each domain, we provide 3 example demonstrations for 2 static, real source sentences. This is inspired by the demonstration aspect of~\citet{he-etal-2024-exploring}, but we elicit 3 demonstration pairs at a time instead of 1.

\paragraph{Best practices} To easily parse the 3 demonstration pairs, we ask for a prescribed JSON output format. We also find that providing static few-shot exemplars of the demonstration task is key to both diversity among the 3 demonstrations, and output format adherence.  We use a different set of exemplars for each domain, drawn from the train set. We perform ablations on the contributions of different aspects of generated demonstrations in \S\ref{sec:ablate_gen_demos}.

\subsubsection{Terminology Generation}
We design a prompt to generate terminologies from a single source sentence (Figure~\ref{fig:prompt_term_gen}), also using 2 static, real sentence pairs for each domain.
This follows in the spirit of the research step of~\citet{briakou2024sbys}, where they explain this as having the LLM perform intermediate reasoning about hard-to-translate parts. However, there are several differences resulting from their focus on document-level MT. We ask generally for terminologies, while they ask specifically for idiomatic expressions, which are more prevalent in long documents. We also prescribe a JSON format (same as for retrieved terms), while theirs allows for free-form output.

\paragraph{Best practices}
We again found that best performance is achieved with static, domain-specific few-shot exemplars of the terminology task, and the prescribed JSON format.

\section{Experimental Setup}
\label{sec:exp}
%We use domain MT as shorthand for the task of domain adaptation for MT. 
\paragraph{Dataset} We experiment with the multi-domain dataset~\cite{aharoni-goldberg-2020-unsupervised}, using the filtered version provided by~\citet{tan-etal-2024-narrowing}, with 3 domains: law, medical, and Koran. Multi-domain covers the German-English (de-en) direction, and consists of dev and test sets, with \texttildelow{}2000 entries per domain, as well as a train set with 1M+ entries.

\paragraph{LLMs} We perform experiments with two LLMs, the open LLM Gemma-2 27B IT~\cite{gemmateam2024gemmaopenmodelsbased}, and the proprietary Gemini 1.5 Pro~\cite{geminiteam2024gemini15unlockingmultimodal}. We thus can investigate which settings, if any, are more effective with the smaller model vs. a much larger model respectively.

% Our primary results consider Gemini, since it can achieve SOTA performance on many tasks, and we wish to study how it can be improved further. We  study Gemma for reproducibility and to address our main research questions on a smaller model.

\paragraph{Evaluation}
%The most straightforward baseline is \textit{zero-shot}, in which we directly ask the LLM to translate from one language to the other.
We perform zero-shot MT as a baseline, and employ the four settings described in \S\ref{sec:domain_knowledge} for comparison: retrieved demonstrations, retrieved terminologies, generated demonstrations, and generated terminologies.
Appendix~\ref{sec:prompts} lists all prompts used in this work.
Following~\citet{vilar-etal-2023-prompting}, we use a neural automated metric, COMET~\cite{rei-etal-2022-comet}. While prior work also considered the lexical metric BLEU, we found that it was overly sensitive to minor rephrasing. This is in line with studies that show neural metrics correlate much better with human judgments of LLM translation quality~\cite{freitag-etal-2021-results, kocmi-etal-2021-ship}.
% domain MT for LLMs primarily considers a lexical metric, BLEU.%\footnote{BLEU scores for comparison are in Appendix X.} 
% We advise that future work in domain-adapted MT follow the general consensus, that only neural automated metrics can properly measure the translation performance of contemporary neural LMs~\cite{freitag-etal-2021-results, kocmi-etal-2021-ship}.

\subsection{Terminology Dictionary Creation}
\label{sec:term_induction}
Our multi-domain test scenario does not come with bilingual terminology dictionaries for its domains. However, we can create them from the provided training split, following the methodology in prior work~\cite{moslem-etal-2023-adaptive, semenov-etal-2023-findings}.\footnote{We did not perform human post-editing due to the datastore's size (700K), but we note in an experiment by ~\citet{moslem-etal-2023-adaptive}, they found humans rated 95\%+ terms as accurate. }
We design a prompt (Figure~\ref{fig:prompt_term_extract}) to extract terminologies from a given source-target text pair, providing 5 static exemplars to demonstrate what is meant by ``terminology''. We then apply this to each pair from the train split. Then, we aggregate all of the output terms, to get one large dictionary with one-to-many mappings.\footnote{For quality controls, we kept only entries where 1) target terms have  >10\% usage and 2) both sides of terms match.} 
%Now, we turn to the source lines from the test split. We iterate over all source-side terms, and if one exists in the source line, we add it to the line's term dictionary (one-to-many).
% For oracle-level, we iterate to find source-side matches, and then additionally iterate to find target-side matches. If both exist, we add the one-to-one source-target mapping to the line's term dictionary.
% We have therefore created two tasks on the same MT dataset: with \textit{global-level} terminologies, and with \textit{oracle-level} terminologies. 
We create a separate global terminology for each of the three domains.%; therefore, we run the process 3 times for the 3 domains of multi-domain. %We release our extra annotations to facilitate future research.\footnote{Link provided after peer review.}

Given the large size of the training split (700K entries), we make two adjustments to reduce the number of model calls. First, we batch five test pairs at a time into a single call. Second, we consider only the subset of train entries that were ever retrieved by BM25 over the test set (i.e. the entries that are actually relevant); this constitutes 70K entries, or about 10\% of the total entries.

%Key is that we apply to a parallel corpus which also contains a data split used for demonstration retrieval, multi-domain. 
Note that the train split is also used for demonstration retrieval, therefore enabling a controlled comparison between the two external knowledge sources. Furthermore, unlike prior work using one-to-one terminology mappings, we explore a more realistic one-to-many scenario, with all possible translations in the prompt for the LLM to select.

%%%%%%%%%
\begin{table*}[t!]
\centering
\scalebox{0.73}{
\begin{tabular}{ll|lc|lc|lc|lc|lc|lc}

\arrayrulecolor{gray!20}
%\toprule
%
\rowcolor{gray!10}
\multicolumn{2}{l}{\textit{\textbf{Domain Knowledge?}}}    & \multicolumn{6}{c}{\texttt{Gemma-2 27B IT}} & \multicolumn{6}{c}{\texttt{Gemini 1.5 Pro}}   \\
\rowcolor{gray!10}
& & \multicolumn{2}{c}{\textbf{Law}} & \multicolumn{2}{c}{\textbf{Med.}} &   \multicolumn{2}{c}{\textbf{Koran}} & \multicolumn{2}{c}{\textbf{Law}} & \multicolumn{2}{c}{\textbf{Med.}} &   \multicolumn{2}{c}{\textbf{Koran}} \\
 \faBan \textbf{\texttt{ zero-shot}} & &  \cboxl{84.8}  &  & \cboxm{85.2}  & &\cboxk{75.1} & & \cboxl{86.6} &  & \cboxm{88.2} & & \cboxk{76.3} & \\
\hline
\multirow{2}{*}{\faStackOverflow \textbf{\texttt{  retrieved}}} & \texttt{terms}  & \cboxlb{85.9} & \good{1.1} & \cboxmb{87.8} & \good{2.6} & \cboxk{74.6} & \bad{0.5} & \cboxlb{86.9} & \good{0.3} & \cboxm{88.5} & \good{0.3} & \cboxk{74.9} & \bad{1.4}\\
%
%\hline
& \texttt{demos} &  \cboxlb{88.6} & \good{3.8} & \cboxmb{89.9} & \good{4.7} & \cboxkb{76.7} & \good{1.6}  & \cboxlb{89.3} & \good{2.7} & \cboxmb{89.9} & \good{1.7} & \cboxk{76.4} & \good{0.1} \\
\hline
\multirow{2}{*}{\faGears  \textbf{\texttt{  generated}}} & \texttt{terms} & \cboxl{85.2} & \good{0.4} & \cboxmb{87.1} & \good{1.9}  & \cboxkb{75.7} & \good{0.6} & \cboxl{86.7} & \good{0.1} & \cboxm{88.1} & \bad{0.1} & \cboxkb{76.9} & \good{0.6} \\
%
%\hline
& \texttt{demos} & \cboxlb{86.0} & \good{1.2} & \cboxmb{88.1} & \good{2.9} & \cboxkb{76.1} & \good{1.0} & \cboxlb{87.2} & \good{0.6}   & \cboxmb{88.8} & \good{0.6} & \cboxkb{76.7} & \good{0.4} \\
%\bottomrule
\end{tabular}}

\caption{Results for MT using the COMET22 metrics, comparing the knowledge sources, retrieved and generated, and the strategies, demonstrations (demos) or terminology (terms). Significant improvements ($p<0.05$) over the zero-shot baseline are marked with *. Demonstrations outperform terminology, and retrieval outperforms generation. Generation is especially effective for the smaller Gemma model. \label{tab:main_results}}
\end{table*}

\section{Results}
\label{sec:results}
Table~\ref{tab:main_results} presents our primary results, comparing LLM translation enhanced with domain-specific knowledge in the form of translation demonstrations or bilingual terminology, with the artifacts derived from either external retrieval (\S\ref{sec:external}) or internal generation (\S\ref{sec:simulate}). First, in line with prior work, we confirm that retrieved demonstrations improve over zero-shot across models and domains studied. We next describe the three main findings.

\paragraph{Demonstrations outperform terminology} For all models and domains studied, knowledge provided in the form of demonstrations consistently outperforms terminology. 
For Gemma, we see that all settings improve performance,\footnote{To explain the outliers for Koran ($-1.4,  -0.5$), our manual analysis found term inconsistency -- high-frequency source terms mapped to multiple, equally-valid target terms.}  but the improvements from demonstrations are markedly larger.
%In fact, demonstration generation (+1.0 to +2.9) outperforms  terminology retrieval (+1.1 to +2.6). 
The differential is more pronounced for Gemini, which starts from a much stronger baseline than Gemma.  Terms, either retrieved or generated, do not provide much of a boost over zero-shot for Gemini, while demonstrations result in significant improvements.  
The takeaway for this finding is that for weaker models, providing domain knowledge from any source or strategy is beneficial. Conversely, stronger models do not benefit from domain-specific terminology, but only from more complete demonstrations of the task. %, and additional hints are unneeded.

% This effect is more pronounced within the stronger Gemini 1.5 Pro model: demonstrations consistently elicit large gains in translation quality over zero-shot, while effects of terminologies are mixed. Retrieved terms show small improvements for law and medical, while generated terms actually underperforms zero-shot for all 3 domains. On the other hand, Gemma-2 27B, which starts from a lower zero-shot baseline compared to Gemini (-1 to -3 COMET), benefits in both strategies of retrieved terms and generated terms.\footnote{For retrieved terms vs. zero-shot, Koran domain is an outlier: $-1.4$ for Gemini, $-0.5$ for Gemma. This is explained by term inconsistency in Koran data, as in our manual analysis of the global terminology, we found that high-frequency source terms often mapped to multiple, equally-valid target terms.} This result highlights that domain knowledge, regardless of its source, is most beneficial for weaker models, and demonstrations are a more effective strategy than terminologies. Conversely, for stronger models (Gemini), which may already possess relevant domain knowledge through pre-training, terminology hints in the prompt are unneeded, and possibly damaging.

\paragraph{Retrieval outperforms generation} The second notable trend across models and domains is that retrieval consistently outperforms generation. With Gemma, demonstration generation outperforms zero-shot by +2.3 (averaged across domains), while retrieval further improves to +3.4. For the more powerful Gemini, the differential is larger -- demonstration generation outperforms zero-shot by +0.5, while retrieval by +1.8.

% The advantage is particularly pronounced when improving a strong base model like Gemini. While generated demonstrations do improve zero-shot translation quality, the gains are modest ($+0.4$ to $+0.6$). In contrast, retrieved demonstrations yield larger improvements, such as $+2.9$ in law and $+2.0$ in medicine. Similar trends are observed with the weaker Gemma-2 model, where generated demonstrations underperform retrieved ones but show larger improvements over zero-shot translations (averaged across domains, $+2.3$ for generation vs. $+3.6$ for retrieval).

% Not sure about this framing.
% \paragraph{Generation of demonstrations efficiently boosts weaker model's translations} 
\paragraph{Generated domain-specific demonstrations boost weaker model's translations}
Taking the prior two findings together, we can bootstrap domain-adapted MT knowledge from an LLM's own parametric memory, with the two-stage approach of first generating demonstrations, then translating. This improvement especially pronounced with Gemma (+2.3 vs. +0.5 over respective zero-shot).
% This bootstrapping only requires two static domain-specific examples, and so is far cheaper than retrieval methods which require large amounts of external resources.
% Generation introduces a minimal amount of external domain knowledge, requiring only two static domain-specific examples. Compare this with the two retrieval methods, which require either hundreds of thousands of translation demonstrations or a large bilingual dictionary of terminology. This makes the effectiveness of generated demonstrations with weaker models become even more impressive.
In fact, this empowers a smaller model (Gemma) to close the gap with a larger model's (Gemini) zero-shot results, as can be seen by comparing, in Table \ref{tab:main_results}, the bottom left and top right rows. The gains in medical (+2.9) and Koran domains (+1.0) result in statistically equivalent scores. Law domain incurs a decent gain (+1.2), but still is below Gemini (86.0 < 86.6). It is worth emphasizing that acquiring extensive resources for novel specialist domains is expensive; but this straightforward approach can be effective.\footnote{Note that the experiment from \S\ref{sec:ablate_gen_demos} shows that domain-specificity is the main contributor, rather than proximity to the current instance.  These generated demos are only as effective as real static, but domain-specific demos.}
% This highlights the efficiency of the generation of intermediate representations as a technique for enhancing weaker models, particularly in scenarios where access to extensive external resources is limited.

\paragraph{Comparisons with Prior Results} 
We can also compare our results with demonstrations to those from the recent study by \citet{tan-etal-2024-narrowing}, who use the \texttt{gpt-3.5-turbo-0301} LLM. Their zero-shot results are most comparable to Gemma's: 84.4, 86.2, 75.1. Their results for retrieved demonstrations are also comparable: 88.2, 89.6, 76.5. The other 3 settings, retrieving terms and both generation ones, are new to our work -- and we re-emphasize here the value of our controlled setting in facilitating fair comparison between them all.

% They  reports gains from $84.4 \rightarrow 88.2$ (law), $86.2 \rightarrow 89.6$ (medical), and $75.1 \rightarrow 76.5$ (Koran). Interestingly, the gains from zero-shot to retrieved demonstrations hold even when prompting a notably stronger baseline, i.e., Gemini 1.5 Pro as seen in Table~\ref{tab:main_results}.
% \input{sections/results_old}

\section{Analysis}
\label{sec:analysis}

Demonstrations (both retrieved and generated) are by far the most effective domain adaptation strategy we explored, providing a large boost to both LLMs. In the following sections, we turn to analyses to understand better where the gains are coming from. We begin by analyzing retrieved demonstrations to disentangle contributions from style vs. terminology (\S\ref{sec:style_vs_term}). Then, we investigate the importance of various in-context learning decisions for  generating demonstrations (\S\ref{sec:ablate_gen_demos}). Finally, we study how generated domain knowledge can be distilled at test-time from larger to smaller models (\S\ref{sec:generated_demost_across_llms}).

\subsection{Retrieved Demonstrations: Contributions from Style vs. Terminology}
\label{sec:style_vs_term}

\begin{figure}[t!]
    \centering
    \includegraphics[width=\linewidth]{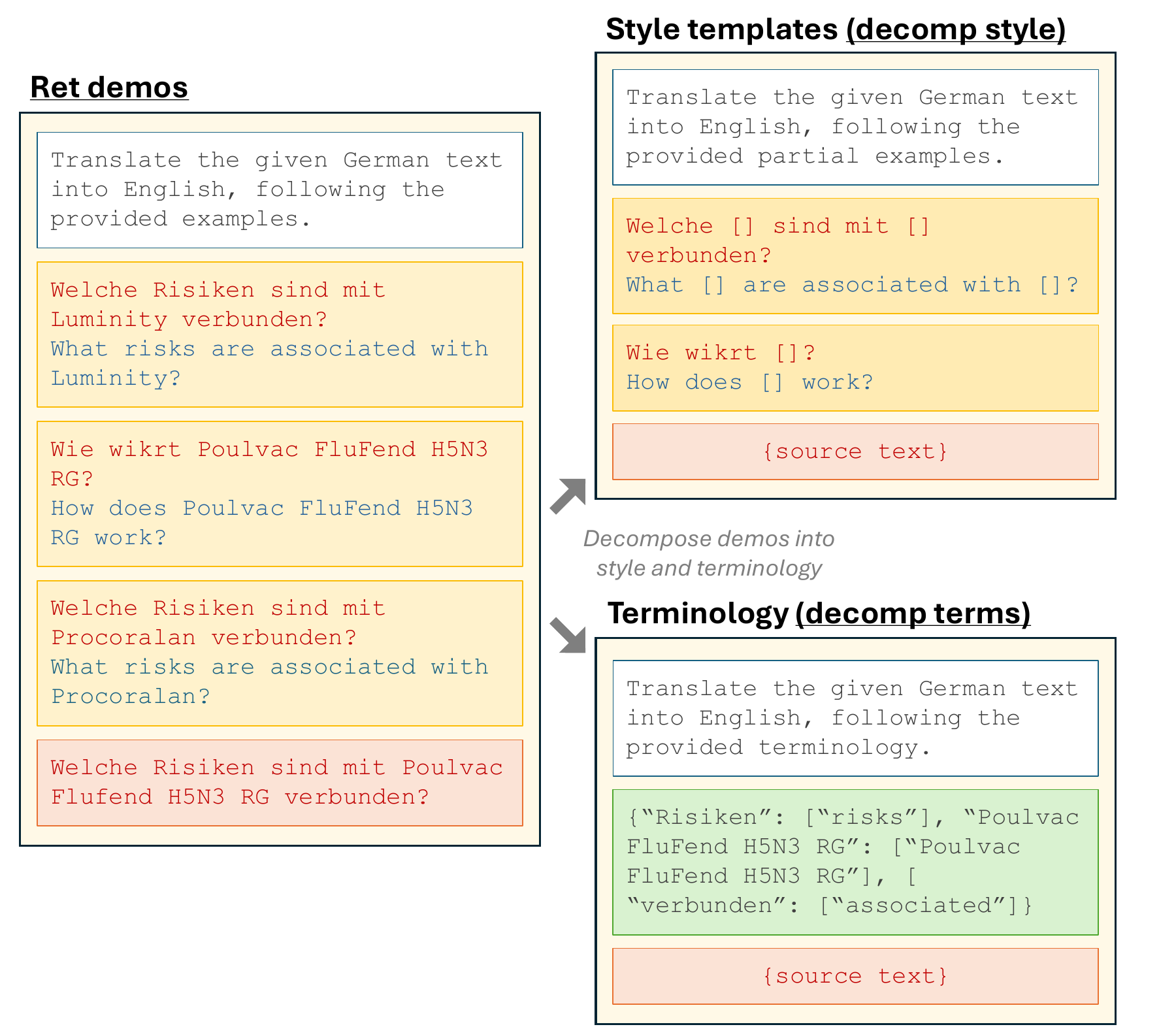}
    \caption{Illustration of our process to decompose the contributions of retrieved demonstrations into style and terminology. We first extract the source-target term pairs using a simple function, and aggregate them into a local terminology. Then, the remaining tokens are the style templates, with the terms masked. Note that in the actual data, we use <MASK> instead of [].}
    \label{fig:decomp}
\end{figure}

What exactly is being conveyed by the retrieved demonstrations? In this section, we take advantage of our careful experimental setup, where our bilingual terminology is derived from the same parallel text used for demonstrations, to disentangle whether demonstrations are more valuable because they assist with proper \textit{terminology} translations in context, or with matching the \textit{style} of the corpus.
%In this following experiment, we disentangle style from terminology in the few-shot retrieved demos, so as to precisely quantify the downstream MT contributions from either.

The core idea behind this experiment is that we can use the same technique to extract bilingual terminology pairs from a translation demonstration (\S\ref{sec:term_induction}), but instead of running it on the whole training corpus, we can run it only on the $k$ demonstrations retrieved to match the current source sentence.
This gives us a \textit{local} terminology, as opposed to a \textit{global} one. Crucially, where the global terminology would present the union of all possible target language translations found throughout the training set for a given source term, the local terminology only presents translations licensed by the $k$ demonstrations. This allows it to take advantage of any disambiguating context in the demonstrations to create more relevant term translations.
%We aggregate all the terms from the retrieved train text pairs to form a local terminology. Filtering to only those terms which appear in the source-side text, we have a \textit{local terminology} to only the information provided by ret demos. For this, we can use the same ret term prompts as before, but with local instead of global terms.

We then define style templates as the inverse -- the remaining tokens, with the bilingual terms on both sides replaced with a \texttt{<MASK>} mask token. For this, we use a similar prompt as for demonstration retrieval, but also  explicitly instruct the LLM to not generate mask tokens in its output (as shown in Figure~\ref{fig:prompt_style}).
Upon manual inspection, these masks appear quite thorough, with most anything that could be considered terminology being masked out.

\paragraph{Results}

\begin{figure}[t!]
    \centering
    \includegraphics[width=\linewidth]{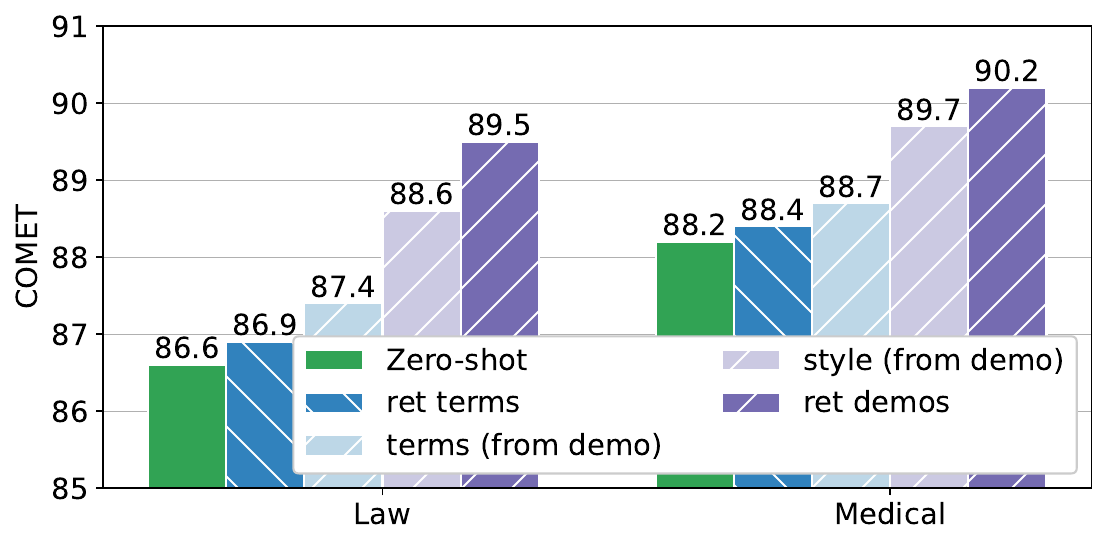}
    \caption{Results for zero-shot, external retrieval, terms from demonstrations, and style from demonstrations.}
    \label{fig:res_decomp}
\end{figure}

We carry out the decomposition experiment using Gemini 1.5 Pro. Figure~\ref{fig:res_decomp} presents our results. We see that compared to zero-shot, using local terms (terms from demonstrations) more than doubles the gains of global terms (retrieved terms). However, style templates (style from demonstrations) further narrow the gap to retrieved demonstrations by 60\% (law) and 75\% (medical).

The combined results from the \textit{terms from demonstrations} and \textit{style from demonstrations} experiments indicate that the primary value from retrieved demonstrations is not contextually appropriate translations of domain-specific terminology. While this is a part of the story, it accounts for only 0.8 (law) and 0.5 (medical) points of the 2.9- and 2.0-point improvements from the retrieved demonstrations. Meanwhile, the \textit{style from demonstrations} scores almost perfectly account for the remainder. This is a strong indicator that the majority of the value of retrieved demonstrations comes from matching the publication style of these corpora, rather than carrying out adaptation to a medical or legal domain. That is, we are doing domain adaptation, but it is to a much more narrow domain than is usually discussed.

%We have thus shown in our decomposition that few-shot demos because 1) they provide specific terminology hints, 2) they provide hints on the desired style, and 3) the contribution from style is more significant than from terms.

These results agree with and reinforce conclusions from recent work. \citet{tan-etal-2024-narrowing} perform a targeted study into translation style, following the same settings as our work -- the multi-domain dataset and a strong proprietary LLM. Their findings between zero-shot and few-shot concur with ours -- that the COMET difference is 2.7, and that zero-shot translations ``have already conveyed the \textit{semantic meaning} of the source sentence, albeit with some variations in \textit{lexical choices} and sentence structure.'' They therefore propose ``style learning,'' which is a method that retrieves related target sentences from a monolingual target corpus. They find that style learning achieves 70\% of the gains of demonstration retrieval. However, by only removing the source side of the demonstrations, the  exemplars still implicitly provide both style and terminology hints. We add to the discussion by providing a precise, alternative definition of ``style'' as anything outside of terminology, which in turn allows us to cleanly decompose the tokens from each demonstration into two subsets, and assign credit accordingly.

\subsection{Ablation on Generated Demonstrations}
\label{sec:ablate_gen_demos}

Of the two generation approaches, doing so for demonstrations is by far the most successful. As described in \S\ref{sec:simulate}, we made several decisions here: using in-context learning (ICL) with two domain-specific examples, where each has three example outputs. The following ablations explore the effects of each decision: \\

\noindent \textbf{No ICL}  Remove all examples of how to generate demonstrations. \\
\noindent \textbf{General ICL} Replace the 2x3 domain-specific examples of demonstrations\footnote{2x3 means there are two example source texts, which are each followed by three related translation examples.}  with the 5x1 general-domain examples from \citet{he-etal-2024-exploring}. \\
\noindent \textbf{Translation ICL} Use the 2x3 domain-specific examples of demonstrations directly as static demonstrations of \textit{translation}. That is, the generation step is dropped, and the translation step sees the same six translation examples every time. This separates the value of the example-specific demonstration generation stage from the value of the static demonstration examples that it uses.

\begin{table}[t!]
    \centering
    \begin{tabular}{llrrr}
    \textbf{row} & \textbf{ablation} & \textbf{law} & \textbf{med.} & \textbf{Koran} \\ \toprule
    1 & zero-shot       & 86.6  & 88.2 & 76.3  \\ \hline
    2 & no ICL          & 83.9  & 83.8 & 75.6  \\
    3 & general ICL     & 86.9  & 88.5 & 76.0  \\
    4 & translation ICL & 87.2  & 89.1 & 76.4  \\ \hline
    5 & generate demos  & 87.2  & 88.8 & 76.7  \\ \bottomrule                 
    \end{tabular}
    \caption{Results for the ablation on in-context learning (ICL) of demonstration generation, using Gemini as the MT model.}
    \label{tab:gen_demo_ablation}
\end{table}

Results for Gemini are shown in Table~\ref{tab:gen_demo_ablation}.\footnote{We also perform this ablation for Gemma, as shown in Appendix Table~\ref{tab:gen_demo_ablation_gemma}, with similar trends to Gemini.} Comparing zero-shot and no ICL (rows 1 \& 2), we see that removing ICL from the demonstration stage is disastrous. Looking at the system outputs, ICL seems to be essential to ensure that the model consistently carries out the task.\footnote{While prompt enginering could have helped, we use the same instructions for fair comparison, noting that part of ICL's value is in making prompt wording less impactful.}
% We likely could have mitigated the worst of the major failures with prompt engineering. Part of the value of ICL is that it makes prompt wording less important.}
Comparing general ICL to generate demos (rows 3 \& 5), we see that roughly half the value of demonstration generation can be retained with general-domain ICL.

However, comparing translation ICL to generate demos (rows 4 \& 5), both achieve similar scores. This result adds insight into two formerly disparate findings. Prior work considering older LLMs, discussed two factors for ICL exemplars: coverage of a domain~\citet{agrawal-etal-2023-context}, and their quality~\cite{vilar-etal-2023-prompting}. Our finding here provides evidence that, for current LLMs with strong zero-shot MT performance, the primary value of ICL is in the domain-specificity, especially in style. This can be equally as validly be obtained from static few-shot exemplars or generated demonstrations.

% the solid scores for translation ICL, which uses domain-specific examples only, without generation, indicates that the domain information from these examples is quite valuable, and demonstration generation is preserving -- rather than enhancing -- that value.

\subsection{Cross-LLM Knowledge Generation}\label{sec:generated_demost_across_llms}
\begin{table}[t]
\centering
\scalebox{0.95}{
\begin{tabular}{lllll}
\textbf{strategy}          & \textbf{gen. LLM} & \textbf{law}  & \textbf{med.} & \textbf{koran} \\
\toprule
\multirow{2}{*}{\faGears \texttt{ demos}}  & Gemma               & 86.0          & 88.1          & 76.1           \\
                           & Gemini              & 86.9* & 88.6* & 76.6*  \\
\midrule
\multirow{2}{*}{\faGears \texttt{ terms}} & Gemma               & 85.2          & 87.1          & 75.7           \\
                           & Gemini              & 85.8*          & 87.5*          & 76.4* \\ \bottomrule
\end{tabular}}
\caption{Results for the ablation on generation-based strategies. \gemmaFull is always used for translation, but the generation model can be either LLM. Significant improvements when using Gemini's generated outputs instead of Gemma's are marked with *.}
\label{tab:cross_llm_ablation}
\end{table}
For the two generation-based settings, the same LLM is used in both the generation stage and the translation stage. To further understand how generation quality affects the final performance, we conduct additional experiments to reuse the generated demonstrations or terminology from Gemini 1.5 Pro to prompt the Gemma 2 27B model for translation. As shown in Table~\ref{tab:cross_llm_ablation}, demonstration generation and terminology generation both benefit greatly from higher quality generations from Gemini, with significant gains in all three domains. 
%For the generated terms, we have mixed results, where only the Koran domain shows significant improvements. 
This shows that higher-quality generated knowledge result in higher-quality translations. The larger Gemini model's knowledge can be effectively distilled to the smaller Gemma model, at inference-time, through its translation demonstrations.

\section{Discussion and Conclusion}
\label{sec:conclusion}
We study the problem of domain adaptation for MT with LLMs, one which intuitively speaking, should be well addressed by prompting-time adaptation. Building upon prior work which injects domain-specific knowledge into prompts, we perform a thorough study into how this knowledge can best be acquired in terms of strategy, demonstrations or terminologies, and sources, retrieval or generation.
% We revisit domain adaptation for MT, and investigate the value of external and internal resources for improving LLM translation. The two knowledge strategies are demonstrations of translations and a global terminology.

Our main study shows that demonstrations outperform terminology, and knowledge retrieval consistently outperforms generation. Furthermore, generation of domain-specific demonstrations can viably improve weaker model's performance, closing the gap with a larger model's zero-shot performance (though comparable to static exemplars). We gain additional insights with our further analyses.
% Our primary findings are that demonstrations outperform terminology, and that generating similar resources by accessing the LLM's parametric memory is especially viable for MT improvements for weaker LLMs.Generation of demos enables a smaller Gemma model to close the gap with the larger Gemini's zero-shot performance.
% . Notably, generated demonstrations on Gemma recover >50\% of the gains from retrieved demonstrations, and this even surpasses the far stronger Gemini with zero-shot MT.
Notably, we explore the connection between the strategies, characterizing demonstrations as providing both terminology hints and style hints. Our decomposition of the contributions of demonstrations finds that the majority of the gains (\texttildelow{}65\%) come from style over terminology. 
% Finally, we perform ablations on MT with generated demonstrations to show our proper design, and the demonstrations' value across LLMs.

Taken together, our work indicates that for the law, medical and Koran domains of the commonly-used multi-domain scenario, large LLMs need very little terminology help, and the improvements from demonstrations are more so from matching corpus style than from better conveying domain-specific semantics.
Our work takes a first step in surfacing the domain-specific knowledge of smaller LLMs through generation, and we look forward to more informed approaches in future work. Meanwhile for the largest LLMs, we recommend as the most promising direction to construct a new MT adaptation scenario that challenges even their broad base of parametric knowledge, perhaps with reference to pretraining cut-off dates.

%Our work presents a careful analysis, with clear motivation on experiments, questions on prior assumptions, and documentation on prompts and design decisions. 
%Traditionally, providing specialist domain knowledge for domain-adapted MT has been addressed with user-defined terminologies. Our findings present two takeaways unlocked by the inference-time flexibility of LLMs: demonstrations may be as, or more, helpful than terminologies, and the rich internal knowledge of LLMs themselves can serve as parametric stores of domain knowledge.

\section*{Limitations}
While our work aims to study the general problem of domain adaptation for machine translation, we studied only a single dataset, which covered 3 domains and 1 language pair. This follows prior work which only studies this one dataset, multi-domain. There are additionally no other suitable datasets for our setting, which we discuss in Appendix~\ref{sec:other_ds}. We noted the limitations of this dataset, in not posing enough of a domain-adaptation challenge for current LLMs. We call on future work to design more up-to-date, comprehensive domain-adapted MT datasets.

For demonstration retrieval, we used only the BM25 algorithm. While this simple approach generally finds relevant demonstrations, prior works have explored more informed ways to do so. We reiterate that improving retrieval-based few-shot MT is not the goal of this work; rather, we aim to understand why it works well, and whether generating from parametric memory alone is viable. Our analysis, including our decomposition of demonstrations into style and terms, can also be applied to demonstrations obtained from any other similarity method.

Our use of a silver terminology built by LLM may lead to an under-estimation of the value of retrieved knowledge from bilingual terminology dictionaries. Likewise, our decomposition of demonstrations into terminology entries and style templates may be affected by the LLM's terminology-extraction errors. As mentioned in the main text, prior work indicates that these techniques (with older LLMs) should be roughly 95\% accurate~\cite{moslem-etal-2023-adaptive}.

We also did not look into the pretraining data, to check if there is data leakage between these datasets, with train/dev/test splits freely available on the web. This is a general concern with all research using proprietary models. However, our work mitigates these concerns through its analytical approach, which contributes an understanding of the task and methods surrounding it, rather than aiming for SOTA numbers on the task. The fact that external demonstration retrieval does improve COMET scores for multi-domain indicates that, at the very least, the paired translations have not been memorized. Also, consider the Koran domain. While an LLM have undoubtedly seen Koran text during training, because there are multiple translations of the Koran into both English and in German, there is no exact 1-1 mapping with respect to the translations used in this dataset.

%However, the fact that gold terminologies do not improve COMET for WMT21 suggests that its covered domain has been entirely internalized in Gemini 1.5 Pro.

% \section{Acknowledgments}
% We thank Markus Freitag ... for their feedback throughout the course of this project.
% We thank Weiting Tan for providing us with the filtered version of the multi-domain test set.

\bibliography{filtered_anthology,custom}
% \bibliographystyle{acl_natbib_from_naacl} % ONLY NEEDED FOR naacl2021.sty
% \bibliography{custom, an}

\newpage
\appendix

\begin{table}[t!]
    \centering
    \begin{tabular}{lrrr}
    \textbf{ablation} & \textbf{law} & \textbf{med.} & \textbf{Koran} \\ \toprule
    zero-shot       & 84.8  & 85.2 & 75.1  \\ \midrule
    no ICL          & 85.2  & 87.5 & 75.1  \\
    general ICL     & 85.7  & 87.8 & 75.6  \\
    translation ICL & 86.3  & 88.2 & 76.3  \\ \midrule
    generate demos  & 86.0  & 88.1 & 76.1  \\ \bottomrule
    \end{tabular}
    \caption{Ablation results for in-context learning (ICL) of demonstration generation. We report comet22 scores using Gemma-2 27B as the model.}
    \label{tab:gen_demo_ablation_gemma}
\end{table}

\section{Other MT with Terminology Datasets}
\label{sec:other_ds}
We did not use the datasets from WMT21 and WMT23 shared tasks on MT with terminologies. They do not include datastore for retrieving demonstrations, as well as each having its own concerns. For WMT21, we found that MT performance for zero-shot and using gold terms was equivalent (87.0 vs. 86.8 COMET22). This is due to contemporary LLM pretraining data containing a lot of COVID domain text, making it no longer a specialist domain. For WMT23, terminologies are internally defined -- i.e., written directly with respect to each test and dev bitext. As we argued earlier, terminologies should be considered as external, pre-defined resources.
We therefore recommend that both WMT21 and WMT23 datasets are outdated with current LLMs, and their use should be avoided.

\citet{aycock-bawden-2024-topic} introduce a domain-adapted MT dataset, which they curate as a subset of existing MT resources from the OPUS project. This covers 7 domains and 11 languages. However, for all domains of their dataset, there is no large-scale data-store for demonstration retrieval; they only perform retrieval -- proposing a topic-model guided exemplar selection method, which they show beats BM25 -- over the very small development splits. Our work therefore considers only the multi-domain dataset, as it widely used for domain-adapted MT, and also satisfies our external resource requirements. 

\section{Prompts Used}
\label{sec:prompts}
We reproduce the exact prompts used below, where \texttt{\{<some\_var>\}} are variables which are filled per prompt, and \texttt{[<some\_ex>]} are the static exemplars which are filled per-domain.

\begin{figure*}[t]
    \centering
    \begin{tcolorbox}[enhanced, width=\linewidth, boxrule=0.8mm] % breakable
        \small
        \begin{verbatim}
Instruction: Translate the following {src_full} text into {tgt_full} and output the result
in JSON format using "translation" as the key.
{source_language_name}: {source_text}
{target_language_name}:
\end{verbatim}
    \end{tcolorbox}
    \centering
    \caption{Prompt for zero-shot MT.}
    \label{fig:prompt_mt_zs}
\end{figure*}

\begin{figure*}[t]
    \centering
    \begin{tcolorbox}[enhanced, width=\linewidth, boxrule=0.8mm] % breakable
        \small
        \begin{verbatim}
You are tasked with translating {source_language_name} to {target_language_name}. You are provided
several example translations, and you should follow their example to translate the given
{source_language_name} sentence.
{demo_examples}
{source_language_name}: {source_text}
{target_language_name}:
\end{verbatim}
    \end{tcolorbox}
    \centering
    \caption{Prompt for MT with \textit{demonstrations} (also known as few-shot MT in prior work). This prompt is used for both demonstration retrieval and demonstration generation.}
    \label{fig:prompt_demos}
\end{figure*}

\begin{figure*}[t]
    \centering
    \begin{tcolorbox}[enhanced, width=\linewidth, boxrule=0.8mm] % breakable
        \small
        \begin{verbatim}
Your task is to translate a piece of text from {source_language_name} into {target_language_name}.
You are provided a list of terminology dictionaries. Each dictionary has a single source term (key
"de"), and multiple candidate translated terms (key "en") -- pick the most appropriate translated
term for the source sentence. Note that the terminologies have lowercased terms, but you should
consider proper casing when translating into {target_language_name}. Based on these terminologies,
output your best one translation.
{examples}
Terminology: {terminology}
{source_language_name}: {source_text}
{target_language_name}:
\end{verbatim}
    \end{tcolorbox}
    \centering
    \caption{Prompt for MT with \textit{terminologies}. This prompt is used for both terminology retrieval and terminology generation.}
    \label{fig:prompt_terms}
\end{figure*}

\begin{figure*}[t]
    \centering
    \begin{tcolorbox}[enhanced, width=\linewidth, boxrule=0.8mm] % breakable
        \small
        \begin{verbatim}
You are tasked with translating {source_language_name} to {target_language_name}. You are provided 
several example translations, and you should follow their example to translate the given 
{source_language_name} sentence. Note that the examples might contain special mask tokens <MASK> but 
in your output, please do not use any such tokens.

[few_shot_examples]
{source_language_name}: {source_text}
{target_language_name}:
\end{verbatim}
    \end{tcolorbox}
    \centering
    \caption{Prompt for MT with \textit{style from demonstrations}. Recall that in this setting, we provide the retrieved demonstrations, but with the terminologies masked out -- i.e., the style contribution is the inverse of the terminology contribution.}
    \label{fig:prompt_style}
\end{figure*}

\begin{figure*}[t]
    \centering
    \begin{tcolorbox}[enhanced, width=\linewidth, boxrule=0.8mm, colback=blue!5, colframe=blue!60] % breakable
        \small
        \begin{verbatim}
You are given a {source_language_name} source text, and asked to write exactly 3 text pairs. A text 
pair consists of a {source_language_name} text, which is related to but different from the source 
text, and its translation into {target_language_name}. You should do your best to ensure that your 
{source_language_name} texts have similar style to the source text. Following the provided examples,
output each pair as a JSON dictionary, with keys "de" and "en". Each dictionary should be on a
separate line.
[demo_examples]
{source_language_name} source: {source_text}
Pair 1:
\end{verbatim}
    \end{tcolorbox}
    \centering
    \caption{Prompt for synthetic \textit{demonstration generation}. \texttt{[demo\_examples]} are static exemplars for this task; see below.}
    \label{fig:prompt_demo_gen}
\end{figure*}

\begin{figure*}[t]
    \centering
    \begin{tcolorbox}[enhanced, width=\linewidth, boxrule=0.8mm, colback=blue!5, colframe=blue!60] % breakable
        \small
        \begin{verbatim}
German source: Die EDGE- und EDGE-II-Studien verglichen die gastrointestinale Verträglichkeit
von Etoricoxib mit der von Diclofenac.
Pair 1: {"de": "Die kardiorenalen Ergebnisse der EDGE- und EDGE-II-Studien entsprachen den
für die MEDAL- Studie beschriebenen.", "en": "The cardiorenal results for EDGE and EDGE II 
were consistent with those described for the MEDAL Study."}
Pair 2: {"de": "Eine langsame Dosissteigerung kann die gastrointestinale Verträglichkeit
ebenfalls verbessern.", "en": "A slow increase in the dose may also improve gastrointestinal 
tolerability."}
Pair 3: {"de": "Die Häufigkeit von unerwünschten Ereignissen in EDGE und EDGE II sowie die 
Häufigkeit von als schwerwiegend erachteten oder zum Studienabbruch führenden unerwünschten 
Ereignissen in der MEDAL-Studie war unter Etoricoxib höher als unter Diclofenac.", "en": "The
incidence of adverse experiences in EDGE and EDGE II and of adverse experiences considered
serious or resulting in discontinuation in the MEDAL study was higher with etoricoxib than
diclofenac."}
---
German source: 3 ml Lösung in einer Patrone aus Glas (Glasart 1), mit einem Kolben (Brombutylgummi)
und einem Stopfen (Brombutylgummi/Polyisopren) in einem Umkarton.
Pair 1: {"de": "3 ml Lösung in einer Patrone aus Glas (Glasart 1), mit einem Kolben (Brombutylgummi)
und einem Stopfen (Brombutylgummi/Polyisopren) in einem Fertigpen (Mehrdosen-Einwegspritze aus
Polypropylen).", "en": "3 ml solution in a cartridge (type 1 glass) with a plunger (bromobutyl) and 
a stopper (bromobutyl/ polyisoprene) contained in a pre-filled pen (multidose disposable pen) 
(polypropylene)."}
Pair 2: {"de": "3 ml Suspension in einer Patrone (farbloses Glas, Typ 1) mit einem Kolben 
(Brombutylgummi, Typ 1) und einer Bördelkappe (Aluminium) mit einem Stopfen (Brombutyl- oder 
Polyisopren- Brombutylgummi, Typ 1).", "en": "3 ml suspension in a cartridge (type 1 colourless 
glass) with a plunger (bromobutyl rubber (type 1)) and a flanged cap (aluminium) with a stopper
(bromobutyl or laminate of polyisoprene and bromobutyl rubber (type 1))."}
Pair 3: {"de": "5 ml Lösung in einer Durchstechflasche (farbloses Glas, Typ 1) mit einer
Bördelkappe (Aluminium), einem Stopfen (Chlorbutylgummi, Typ 1) und einem Abreißdeckel 
(Polypropylen).", "en": "5 ml solution in a vial (type 1 colourless glass) with a flanged cap
(aluminium), a stopper (chlorobutyl rubber (type 1)) and a tear-off cap (polypropylene)."}
---
\end{verbatim}
    \end{tcolorbox}
    \centering
    \caption{Static 2-shot exemplars used for the synthetic \textit{demonstration generation} prompt (Figure~\ref{fig:prompt_demo_gen}). Each exemplar has 3 output sentences. Here we show the exemplars for the medical domain.}
    \label{fig:demo_demo_medical}
\end{figure*}

\begin{figure*}[t]
    \centering
    \begin{tcolorbox}[enhanced, width=\linewidth, boxrule=0.8mm, colback=red!5, colframe=red!60] % breakable
        \small
        \begin{verbatim}
You are given a {source_language_name} source text, and asked to extract a bilingual terminology
that translates key terms from the source text into {target_language_name}. Each entry in the
terminology should have a {source_language_name} term and a list of possible {target_language_name}
translations. Following the provided examples, output each pair as a JSON dictionary, with keys "de"
and "en". Each dictionary should be on a separate line.
[term_examples]
{source_language_name} source: {source_text}
Term 1:
\end{verbatim}
    \end{tcolorbox}
    \centering
    \caption{Prompt for synthetic \textit{terminology generation}. \texttt{[term\_examples]} are static exemplars for this task; see below.}
    \label{fig:prompt_term_gen}
\end{figure*}

\begin{figure*}[t]
    \centering
    \begin{tcolorbox}[enhanced, width=\linewidth, boxrule=0.8mm, colback=red!5, colframe=red!60] % breakable
        \small
        \begin{verbatim}
German source: (6) Die Kommission unterrichtete den Antragsteller, andere Gemeinschaftshersteller,
die ausführenden Hersteller in der VR China und in den USA, bekanntermaßen betroffene Einführer
und Verwender sowie die Vertreter der Regierungen der VR China und der USA offiziell über die
Einleitung des Verfahrens.
Term 1: {"de": "einleitung des verfahrens", "en": ["initiation of the proceeding",
"opening of the proceedings"]}
Term 2: {"de": "ausführenden hersteller", "en": ["exporting producers"]}
Term 3: {"de": "gemeinschaftshersteller", "en": ["community producers"]}
Term 4: {"de": "antragsteller", "en": ["complainant"]}
Term 5: {"de": "kommission", "en": ["commission"]}
Term 6: {"de": "verfahrens", "en": ["investigation", "procedure"]}
Term 7: {"de": "einführer", "en": ["importers"]}
Term 8: {"de": "verwender", "en": ["users"]}
Term 9: {"de": "vertreter", "en": ["representatives"]}
Term 10: {"de": "vr china", "en": ["prc"]}
---
German source: ENTSCHEIDUNG DER KOMMISSION vom 25. Februar 1998 zum Fragebogen für die Berichte der
Mitgliedstaaten über die Umsetzung der Richtlinie 94/67/EG des Rates über die Verbrennung 
gefährlicher Abfälle (Umsetzung der Richtlinie 91/692/EWG des Rates) (Text von Bedeutung 
für den EWR) (98/184/EG)
Term 1: {"de": "verbrennung gefährlicher abfälle", "en": ["incineration of hazardous waste"]}
Term 2: {"de": "fragebogen", "en": ["questionnaire"]}
Term 3: {"de": "richtlinie", "en": ["directive", "guideline"]}
Term 4: {"de": "ewr", "en": ["eea relevance"]}
---

\end{verbatim}
    \end{tcolorbox}
    \centering
    \caption{The static 2-shot exemplars used for the synthetic \textit{terminology generation} prompt (Figure~\ref{fig:prompt_term_gen}). Here we show the exemplars for the law domain.}
    \label{fig:demo_term_law}
\end{figure*}

\begin{figure*}[t]
    \centering
    \begin{tcolorbox}[enhanced, width=\linewidth, boxrule=0.8mm, colback=green!5, colframe=green!60] % breakable
        \small
        \begin{verbatim}
Identify and annotate all terminology entities (consider only consecutive words) from the source
sentences and match them with the counterpart in the target sentences. Your response should follow
the format of the provided examples, so that each numbered source and target pair corresponds to
exactly one terminology line in your response.
[source_examples]
{source_texts}
---
[target_examples]
{target_texts}
---
[term_examples]
\end{verbatim}
    \end{tcolorbox}
    \centering
    \caption{Prompt for \textit{terminology extraction} from source-target text pairs. For each prompt, we batch together 5 text pairs to extract from at a time. \texttt{[source\_examples], [target\_examples], [term\_examples]} are static exemplars for this task; see below.}
    \label{fig:prompt_term_extract}
\end{figure*}

\begin{figure*}[t]
    \centering
    \begin{tcolorbox}[enhanced, width=\linewidth, boxrule=0.8mm, colback=green!5, colframe=green!60] % breakable
        \small
        \begin{verbatim}
source 1: Sag: "Wer hat denn die Schrift hinabgesandt, mit der Musa als Licht und als Rechtleitung
für die Menschen kam?
source 2: Sollte Seine Peinigung über euch nachts oder am Tage hereinbrechen, was wollen denn die
schwer Verfehlenden davon beschleunigen?"
source 3: Unser Herr! Du bist wahrlich Gütig, Barmherzig."
source 4: Und diejenigen, die an Allah und Seine Gesandten glauben, sind die Wahrhaftigen und die
Bezeugenden vor ihrem Herrn; sie werden ihren Lohn und ihr Licht empfangen.
source 5: "Wer sich im Irrtum befindet, dem soll Der Allgnade Erweisende noch mehr davon gewähren!"
Wenn sie dann sehen, was ihnen angedroht wurde: entweder die Peinigung oder die Stunde, dann werden
sie wissen, wer über die schlimmere Stellung und die schwächere Streitmacht verfügt.\end{verbatim}
\tcbline\begin{verbatim}
target 1: Say: "Who sent down the Book that Moses brought as a light and a guidance to men?
target 2: If His chastisement comes upon you by night or day, what part of it will the sinners seek
to hasten?
target 3: Our Lord, surely Thou art the All-gentle, the All-compassionate."
target 4: Those who believe in God and His apostles are true of word and deed; and by their Lord are
considered testifiers of the truth. They have their guerdon and their light.
target 5: "Ar-Rahman extends the life of those who are astray until they come to realise what had 
been promised them was either (physical) affliction or (the terror) of Resurrection. Then will they
know who is worse in position, and who is weak in supporters.\end{verbatim}
\tcbline\begin{verbatim}
terminology 1: [{{"en": "Book", "de": "Schrift"}}, {{"en": "guidance", "de": "Rechtleitung"}}, 
{{"en": "Moses", "de": "Musa"}}]
terminology 2: [{{"en": "chastisement", "de": "Peinigung"}}, {{"en": "sinners", "de": "schwer
Verfehlenden"}}]
terminology 3: [{{"en": "Our Lord", "de": "Unser Herr"}}, {{"en": "All-gentle", "de": "Gütig"}}, 
{{"en": "All-compassionate", "de": "Barmherzig"}}]
terminology 4: [{{"en": "His apostles", "de": "Seine Gesandten"}} {{"en": "true of word and deed",
"de": "die Wahrhaftigen und die Bezeugenden"}}, {{"en": "by their Lord", "de": "vor ihrem Herrn"}},
{{"en": "their guerdon", "de": "ihren Lohn"}}, {{"en": "their light", "de": "ihr Licht"}}]
terminology 5: [{{"en": "Ar-Rahman", "de": "Der Allgnade Erweisende"}}, {{"en": "extends the life",
"de": "noch mehr davon gewähren"}}, {{"en": "those who are astray", "de": "Wer sich im Irrtum
befindet"}}, {{"en": "come to realise", "de": "sehen"}}, {{"en": "promised", "de": "angedroht"}},
{{"en": "(physical) affliction", "de": "Peinigung"}}, {{"en": "(the terror) of Resurrection", "de":
"Stunde"}}, {{"en": "worse in position", "de": "über die schlimmere Stellung"}}, {{"en": "weak in
supporters", "de": "die schwächere Streitmacht"}}]

\end{verbatim}
    \end{tcolorbox}
    \centering
    \caption{Static 5-shot exemplars used for the synthetic \textit{terminology extraction} prompt (Figure~\ref{fig:prompt_term_extract}). We found that this format, where each of the 3 blocks (source, target, terms) are consecutive to each other, gave the most parseable output. Note that the 5 exemplars is the same size as the batches of 5 to extract terminologies from. Here we show the exemplars for the koran domain.}
    \label{fig:demo_term_extract}
\end{figure*}

\end{document}